\documentclass[10pt,twocolumn,letterpaper]{article}

\usepackage{cvpr}              

\definecolor{wine}{RGB}{153,73,70}

\newcommand*{\TitleMethod}{\textcolor{wine}{LaFiTe}\xspace}
\newcommand*{\Method}{{LaFiTe}\xspace}

\definecolor{cvprblue}{rgb}{0.21,0.49,0.74}
\usepackage[pagebackref,breaklinks,colorlinks,allcolors=cvprblue]{hyperref}
\usepackage[table]{xcolor}
\usepackage{multirow}
\usepackage{multicol}

\def\paperID{}
\def\confName{}
\def\confYear{}

\title{\TitleMethod: A Generative \textcolor{wine}{La}tent \textcolor{wine}{Fi}eld for 3D Native \textcolor{wine}{Te}xturing}

\author{
Chia-Hao Chen$^1$ \quad
Zi-Xin Zou$^2$ \quad
Yan-Pei Cao$^2$ \\
Ze Yuan$^3$ \quad
Guan Luo$^1$ \quad
Xiaojuan Qi$^3$ \quad
Ding Liang$^2$ \quad
Song-Hai Zhang$^{1\dagger}$ \quad
Yuan-Chen Guo$^{2\dagger}$ \\[1em]
$^1$Tsinghua University \quad
$^2$VAST \quad
$^3$The University of Hong Kong
}

\begin{document}

\twocolumn[{
\renewcommand\twocolumn[1][]{#1} %
\maketitle
\includegraphics[width=\linewidth]{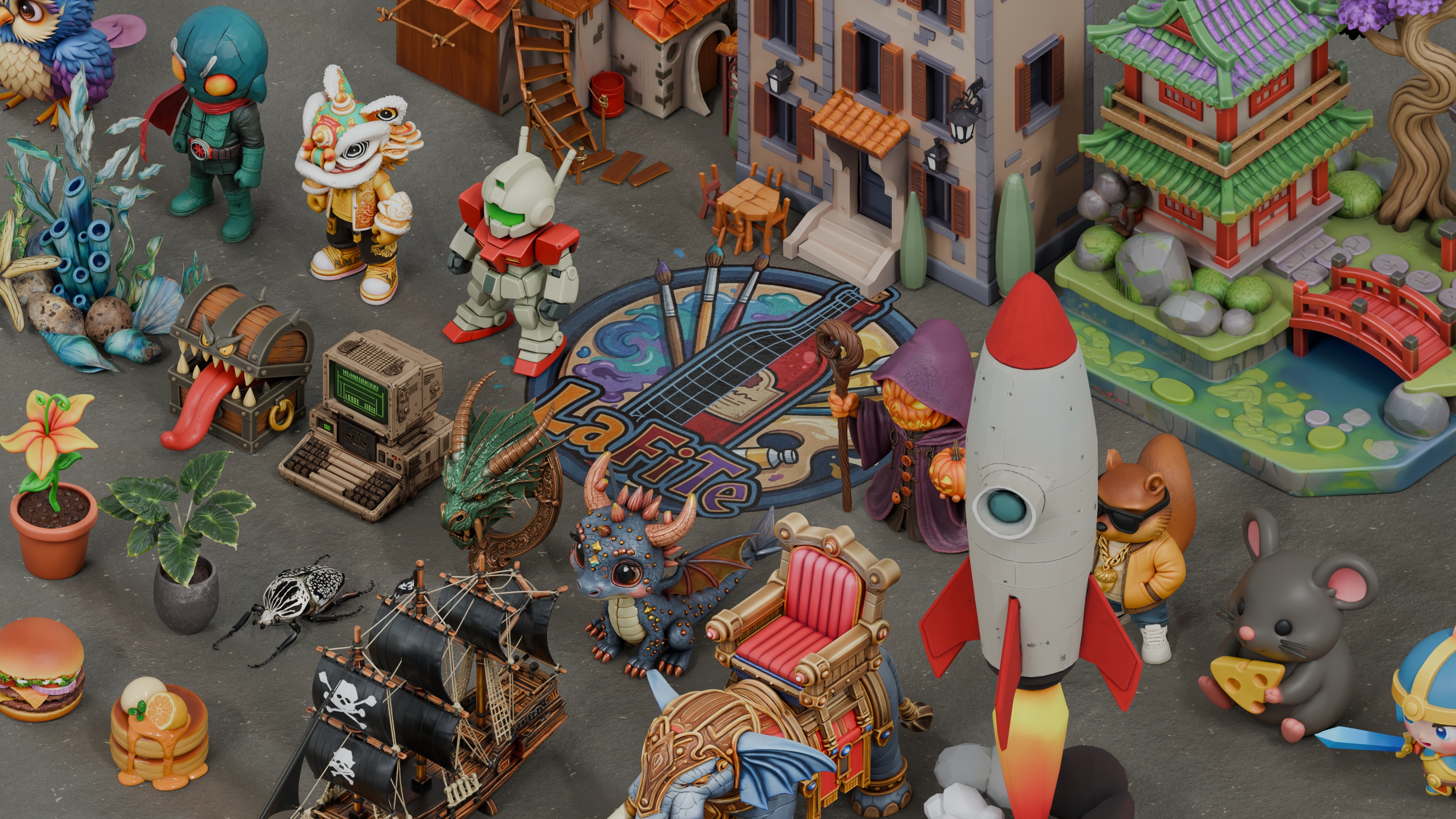}
\captionof{figure}{A gallery of diverse 3D assets textured by our 3D-native framework, \Method, demonstrating \textbf{high-fidelity, seamless textures across a wide range of visual styles}.}
\vspace{4mm}
\label{fig:teaser}

}]

\makeatletter
\def\@fnsymbol#1{\ensuremath{\ifcase#1\or \dagger\or \ddagger\or
   \mathsection\or \mathparagraph\or \|\or **\or \dagger\dagger
   \or \ddagger\ddagger \else\@ctrerr\fi}}
\makeatother
\maketitle
\begingroup
\renewcommand\thefootnote{}      
\footnotetext{$\dagger$ Corresponding authors}
\endgroup

\begin{abstract}
Generating high-fidelity, seamless textures directly on 3D surfaces, what we term 3D-native texturing, remains a fundamental open challenge, with the potential to overcome long-standing limitations of UV-based and multi-view projection methods. However, existing native approaches are constrained by the absence of a powerful and versatile latent representation, which severely limits the fidelity and generality of their generated textures. We identify this representation gap as the principal barrier to further progress.
We introduce \Method, a framework that addresses this challenge by learning to generate textures as a 3D generative sparse latent color field. At its core, LaFiTe employs a variational autoencoder (VAE) to encode complex surface appearance into a sparse, structured latent space, which is subsequently decoded into a continuous color field. This representation achieves unprecedented fidelity, exceeding state-of-the-art methods by \textbf{$>$10 dB PSNR} in reconstruction, by effectively disentangling texture appearance from mesh topology and UV parameterization. Building upon this strong representation, a conditional rectified-flow model synthesizes high-quality, coherent textures across diverse styles and geometries.
Extensive experiments demonstrate that \Method not only sets a new benchmark for 3D-native texturing but also enables flexible downstream applications such as material synthesis and texture super-resolution, paving the way for the next generation of 3D content creation workflows.
\end{abstract}

\section{Introduction}
\label{sec:intro}

The automated generation of high-quality 3D content has become a cornerstone of modern computer graphics and vision. While significant strides have been made in generating complex 3D geometry~\cite{zhang20233dshape2vecset,xiang2025structured}, the creation of detailed, seamless, and coherent textures remains a major bottleneck. Textures often contribute more to the visual richness and realism of an asset than the geometry itself. As such, developing powerful generative models for texturing is a critical and defining challenge for the next generation of AI-driven content creation.

Current approaches for texture generation predominantly fall into two categories, both of which are inherited from 2D-centric paradigms and suffer from fundamental limitations. The most common approach involves generating multiple 2D images of an object from different viewpoints and then projecting them back onto the mesh surface~\cite{oechsle2019texture,chen2023text2tex,liu2024text,he2025materialmvp,liang2025unitex}. This ``multi-view projection'' strategy is inherently flawed - it struggles to reconcile inconsistent predictions caused by cross-view misalignment, occlusions, and view-dependent lighting, inevitably leading to conspicuous seams and artifacts that require complex post-processing to fix. An alternative approach is to generate textures directly in a 2D UV-unwrapped space~\cite{yu2024texgen}. However, this method trades one set of problems for another. It becomes critically dependent on a mesh's UV parameterization, which is non-unique and often introduces severe distortions, leading to textures that are warped or contain seams along UV island boundaries. These approaches are symptomatic treatments, not a cure for the underlying challenge of creating truly 3D-native appearance.

We argue that the only way to guarantee spatial coherence and seamlessness is to generate textures \textit{directly in 3D space}. This native approach sidesteps the inherent contradictions of 2D-based methods. If this path is so clearly superior, why has it not yet dominated? The central roadblock, and the primary focus of this work, has been the lack of a suitable \textbf{3D texture representation}. An ideal representation for generative modeling must satisfy three criteria: it must be \textbf{expressive} enough to capture high-frequency details; it must be \textbf{decoupled} from the mesh's specific topology and UV coordinates; and it must be \textbf{compact and structured} enough to be learned effectively by a generative model.

{To address this critical representation gap, we introduce \textbf{\Method}, a framework built on the central insight that a powerful generative model of 3D textures must rely on a powerful representation. We propose to model 3D texture as a \textbf{sparse latent color field}, and design a novel variational autoencoder (VAE) to learn this color field from data. Our VAE encodes the appearance of a 3D asset, represented as a dense colored point cloud, into a sparse, structured latent feature. This representation is powerful because it is \textbf{local and continuous}; it concentrates modeling capacity near the object's surface and can be queried at any point in space to reconstruct fine-grained details. This fundamentally disentangles texture from the discrete, and often problematic, mesh connectivity and UV coordinates.}

{Furthermore, \Method introduces a second key insight into the geometry conditioning of the generative process. We found that the geometry of the mesh is implicitly but accurately captured by our VAE encoder. By encoding an untextured (``monochrome'') version of the mesh, we can extract a geometry latent that provides a complete, occlusion-free 3D shape guidance. This elegant maneuver allows us to condition our generative model on a perfect 3D geometric signal that is inherently aligned with the texture representation, without the need for a separate geometry encoder. This synergistic design of representation and conditioning is the key to \Method's ability to synthesize textures with state-of-the-art quality and coherence.}

Our main contributions are summarized as follows.
\begin{itemize}

    \item \textbf{A High-Fidelity Latent Color Field Representation:} We are the first to propose and learn a latent color field for native texturing, achieving state-of-the-art reconstruction fidelity (\textbf{$>$10 dB PSNR} improvement over prior work) while being fully decoupled from the mesh topology.

    \item \textbf{A Unified Texture Generation Framework:} We achieve 3D texturing by generating the latent color field using flow matching, and introduce a highly effective method for geometry conditioning by reusing our texture encoder to encode a monochrome point cloud.
    \item \textbf{State-of-the-Art Native Texturing:} \Method significantly outperforms all existing native and projection-based methods, delivering seamless, detailed, and coherent textures across a diverse range of complex 3D assets.

\end{itemize}

\section{Related Work}
\label{sec:rw}

\paragraph{3D Representations and Generative Models.}
The field of 3D content generation has rapidly evolved, driven by advances in both underlying 3D representations and the neural models that create them. Research has explored a wide range of representations, from implicit fields like SDFs~\cite{chen2019implicit,park2019deepsdf} and UDFs~\cite{guillard2022udf, park2019deepsdf} to explicit structures like meshes~\cite{nash2020polygen, siddiqui2024meshgpt, chen2024meshanything}, point clouds, and 3D Gaussians~\cite{kerbl20233d}. The generative models and methods themselves have progressed from score-based distillation~\cite{poole2023dreamfusion,wang2023score,wang2023prolificdreamer,yu2023text,liang2024luciddreamer,guo2023threestudio} on differentiable representations to powerful 3D-native models. A key component in many modern systems is the intermediate neural representation, such as triplanes~\cite{chan2022efficient,hong2024lrm,li2024instantd,zou2024triplane,zhang2024gs}, sparse latents~\cite{xiang2025structured,he2025sparseflex,wu2025direct3d}, or VecSets~\cite{zhang20233dshape2vecset,li2025triposg,zhang2024clay,hunyuan3d_tencent}, which encode 3D information for the generative model. While this body of work provides the foundation, it has largely focused on geometry; our work tackles the distinct challenge of a generative representation for 3D texture.

\paragraph{Multi-View Image Generation.}
A recent common paradigm for 3D texturing is to synthesize multiple 2D views and project them onto a mesh. Initial works, enabled by powerful 2D diffusion models, faced a trade-off between slow, per-object optimization~\cite{guo2023decorate3d} and progressive generation~\cite{richardson2023texture,cao2023texfusion,chen2023text2tex} that risked view misalignment. Subsequent research has therefore focused on improving multi-view consistency~\cite{liu2024text,cheng2025mvpaint,gao2024genesistex,lu2025genesistex2}, either by synchronizing the denoising process across views or by jointly generating all views~\cite{huang2025mv,yuan2025seqtex,yan2025flexpainter,he2025materialmvp,liang2025unitex,bensadoun2024meta,zhang2024clay}. Orthogonal to consistency, a major thrust has been generating physically-based rendering (PBR) materials~\cite{siddiqui2024meta,zhu2024mcmat,zhang2024dreammat,hong2025supermat,huang2025material,he2025materialmvp,birsak2025matclip,fang2024make,chen2023fantasia3d,yeh2024texturedreamer,munkberg2025videomat} to support realistic relighting~\cite{zeng2024rgb,kocsis2025intrinsix,hong2024supermat}. While increasingly sophisticated, these methods all grapple with the fundamental ill-posedness of reconciling multiple 2D images into a single, coherent 3D surface, a problem our native approach sidesteps entirely.

\paragraph{Native Texture Generation.}
To circumvent the inherent issues of 2D projection, native approaches generate texture directly in the object's own space. Some methods operate in a 2D UV-unwrapped space~\cite{yu2023texture,yu2024texgen,yuan2025seqtex}, but this makes them dependent on a non-unique and often distorted parameterization. A more robust direction, and the one we pursue, is to represent textures natively in 3D. Prior works have explored this by learning on mesh surfaces~\cite{siddiqui2022texturify}, 3D texture fields~\cite{oechsle2019texture,xie2024styletex,liang2025unitex,xiang2025structured,huo2024texgen}, and generating colored point clouds or 3DGS~\cite{yu2023texture,liu2024texoct,xiong2025texgaussian} to define spatial texture. However, these pioneering native works have been limited by their chosen representations, which can struggle to capture high-frequency detail within a compact latent space suitable for high-fidelity generative modeling. 

\section{Method}

\begin{figure*}[htbp]
    \centering
    \includegraphics[width=\linewidth,trim=00 00 00 00,clip]{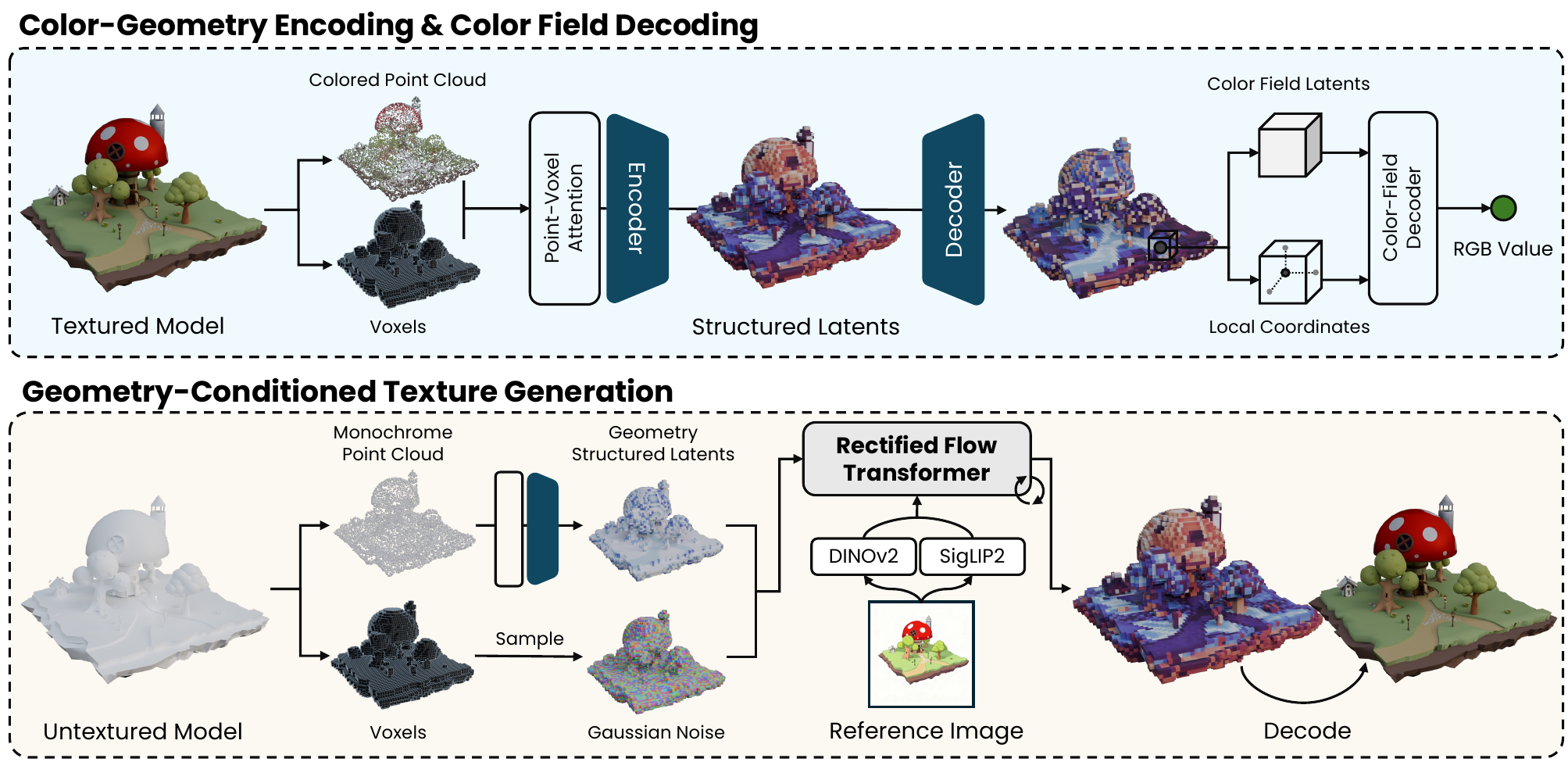}
    \caption{\textbf{The \Method Pipeline.}
     \textbf{(Top)} A VAE autoencoder learns a sparse latent color field representation by encoding a colored point cloud sampled from the mesh surface. \textbf{(Bottom)} For generation, the VAE encoder first extracts a geometry latent from a monochrome point cloud. This geometry latent then conditions a rectified flow model to synthesize a texture latent, which is decoded to the 3D texture.}

    \label{fig:pieline}
\end{figure*}

\label{sec:method}

We introduce \Method, a generative framework for seamless, high-fidelity native texturing on arbitrary mesh topologies. At its core is a novel latent representation of 3D texture, which we detail along with its encode-decode architecture (Sec.~\ref{sec:method-representation}). We then describe our conditional generation model (Sec.~\ref{sec:method-generation}), various downstream applications (Sec.~\ref{sec:method-applications}), and our data curation pipeline (Sec.~\ref{sec:data}).

\subsection{{The \Method Representation: A VAE for Sparse Latent Color Field}}
\label{sec:method-representation}
Unlike dense representations such as triplanes~\cite{chan2022efficient,shue20233d} or dense grids~\cite{mueller2022instant}, we adopt a more efficient sparse structure that has been widely used in recent advanced models~\cite{xiang2025structured,he2025sparseflex,li2025sparc3d}. Specifically, we discretize 3D space into a voxel grid, storing information only in active voxels near the object's surface. Within each active voxel, a continuous implicit function defines the local color field. 
This hybrid approach efficiently concentrates modeling capacity and memory, handles arbitrary topology, and enables the querying of continuous, high-frequency details. To learn this representation, we design a variational autoencoder (VAE) that maps a densely sampled colored points to a compact, sparse-structured latent space, from which the continuous color field is subsequently decoded.

\noindent\textbf{Texture Encoding.}
Recent 3D generative models~\cite{xiang2025structured} {typically achieve this by projecting}
pre-trained 2D features (e.g., DINO~\cite{oquab2023dinov2,simeoni2025dinov3}) from
multiple camera views onto the {on-surface sparse voxels.}
While {effective}, this strategy {exhibits} {three} fundamental limitations: 
(1) The projection process is view-dependent, causing inconsistencies or ambiguity in occluded regions; (2) Pretrained 2D features have limited resolution, restricting high-frequency surface details; (3) Features are tied to the 2D foundation knowledge, which may not optimally represent fine-grained 3D surface attributes such as PBR materials.

In contrast, we propose constructing a representation that directly and unbiasedly captures the high-fidelity texture of a 3D asset by using a colored point cloud $\mathcal{P}=\{x_i=\left(p_i,n_i,c_i\right)\}$ densely sampled from the textured mesh.
Specifically, we uniformly sample points $p_i$ together with their corresponding normals $n_i$ and raw colors $c_i$, which serve as the input to the texture encoder.
This provides the encoder with {native}, unoccluded, and view-independent surface information, minimizing information loss and forming a more accurate ground truth for representation learning.

Given this point cloud, we adopt an encoder architecture similar to SparseFlex~\cite{he2025sparseflex}, but replacing the PointNet~\cite{qi2017pointnet} with a specialized point-voxel attention mechanism for robustly fusing point features within each voxel.
This encoder compresses the detailed, high-frequency texture information into sparse, structured latent features $\mathcal{Z} = \{z_k\}$, where $k$ indexes the set of $L$ on-surface voxels.
The point-voxel attention mechanism comprises an intra-voxel self-attention module and a point-voxel cross-attention module.

The intra-voxel self-attention module processes all points within each voxel, allowing them to interact and aggregate local geometric and appearance features into a set of point-wise features $\tilde{x}$:
\begin{align}
    \tilde{x}_i=\sum_{j=1}^{n}\text{softmax}(\frac{Q_{x_i}K_{x_j}^T}{\sqrt{d}})\cdot V_{x_j},
\end{align}
where $x_j$ denotes the initial point features within the same voxel as $x_i$.

The point-voxel cross-attention module then consolidates these detailed point-wise features into voxel-wise features
$\tilde{v}_k$, representing the encoded texture information for each voxel.
This is achieved by attending from a shared learnable voxel feature $v_k$ to all point-wise features within the voxel:
\begin{align}
    \tilde{v}_k=\sum_{i=1}^{n}\text{softmax}(\frac{Q_{v_k}K_{\tilde{x}_i}^T}{\sqrt{d}})\cdot V_{\tilde{x}_i}.
\end{align}
This design ensures robust capturing the local surface appearance than the average pooling in PointNet. 
The resulting set of voxel embeddings $\{\tilde{v}_k\}$ is then passed through a standard sparse-voxel-based VAE encoder~\cite{xiang2025structured,he2025sparseflex} with shifted window attention
to produce the final
latent {codes} $\{z_k\}$. The overall {texture} encoding process is as follows:
\begin{equation}
    \mathcal{E}: \{\{{x_j}\}_{j=1}^{N_i}\}_{i=1}^L \rightarrow \{z_k\}_{k=1}^L.
\end{equation}
\noindent\textbf{Implicit Geometry Encoding.} 
A crucial property of our encoder is that, besides encoding texture, it naturally captures high-fidelity geometry from the point cloud positions at no extra cost. We exploit this by introducing a \textit{pure geometry latent}: passing a monochrome point cloud -- setting all input colors $c_i$ to a canonical value like white $(1,1,1)$ -- we effectively factor out texture information. 
The resulting geometry latent $z_\text{geo}$ represents only the object's shape. This elegant approach provides an occlusion-free 3D geometric prior for our generative model without a separate geometry network, while ensuring perfect alignment between the shape and texture conditions during synthesis.

\noindent\textbf{Texture Decoding and Supervision.} The decoder's role is to reconstruct the continuous color field $\mathcal{C}(p)$ from the sparse{, structured} latent {representation} $\{z_k\}$.
For any 3D query point $p$ within a voxel $k$, a latent decoder $\mathcal{D}_S$ first maps the corresponding latent $z_k$ to a {higher-resolution} local feature grid $f_k$.
{This local feature grid forms a sparse local color field, from which
the color decoder $\mathcal{D}_F$}
{takes} features interpolated using
the local coordinates of $p$ and outputs the final RGB color $c$ {through a shallow MLP}. The overall VAE decoder is as follows:
\begin{equation}
    \mathcal{D}: \{z_k\}_{k=1}^L \times p_j \rightarrow c_j.
\end{equation}
We train the VAE end-to-end using a reconstruction loss combined with a KL divergence regularization term. The reconstruction loss is a simple L1 loss on the color values of sampled points, which we find effectively promotes sharp details. To encourage the model to learn a robust representation of high-frequency textures, we augment input colors with minor Gaussian noise during training. The overall VAE objective is given by:
\begin{equation}
    \mathcal{L}=\mathbb{E}_{x_i\sim\mathcal{M}}[|\mathcal{D}(\mathcal{E}(\{\hat{x}_i\}), p_j)-\hat{c}_j|] + \mathcal{L}_{\text{KL}}.
\end{equation}
where $\hat{x}_i=(p_i, n_i, \hat{c}_i)$, $\hat{c}_i=c_i+\epsilon$ is the augmented color. Importantly, we do not rely on rendering-based losses (like LPIPS or SSIM), which can introduce blurriness and bias. 
{Instead,} supervision is performed directly in the 3D space, ensuring the learned representation faithfully models the underlying surface texture.

\subsection{Conditional Texture Synthesis via Rectified Flow}
\label{sec:method-generation}
Having established a powerful latent representation for 3D textures, we now detail the generative process for synthesizing novel textures within this space. We employ a conditional rectified flow~\cite{liu2022rectified,labs2025flux1kontextflowmatching} model, which learns to generate texture latents $z$ conditioned on the geometry of the input mesh and a reference image.

\noindent\textbf{Geometry-Conditioned Albedo Generation.}
The cornerstone of our synthesis pipeline is the generation of a base color (albedo) texture. The quality of this generation depends critically on the fidelity of the geometric conditioning. Unlike methods that rely on 2D geometric projections (e.g., position or normal maps), which suffer from occlusions and cross-view inconsistencies, our approach conditions directly on the complete and unambiguous 3D shape.

Specifically, we use the pure geometry latent $z_{\text{geo}}$, derived as described in Sec.~\ref{sec:method-representation}, as the primary condition. This provides the generative model with a {detailed}, occlusion-free
3D geometry prior that is perfectly aligned with the target latent space. The model is trained using the conditional flow matching objective:
\begin{equation}
    \mathcal{L}_\text{albedo}=\mathbb{E}\Vert v(x_t;t, z_\text{geo})-(\epsilon-x_0)\Vert,
\end{equation}
where $c_\text{geo}$ is the geometry condition represented by $\mathcal{E}(\{p_i, n_i, \mathbf{1}\})$, and $x_t$ is the noisy latent in timestep $t$, $x_t=(1-t)x_0+t\epsilon$. 
We adopt a progressive training strategy: The model is first trained on latents at $64^3$ resolution, and then fine-tuned on $128^3$ resolution latents, which are encoded from a denser point cloud sampling.

\noindent\textbf{Disentangled PBR Material Generation.}
Our framework naturally extends to generating physically-based rendering (PBR) materials by learning to synthesize roughness and metallic maps. A key insight is that these material properties are often physically correlated with the underlying albedo; therefore, conditioning solely on geometry is suboptimal. We leverage this insight to create a disentangled, hierarchical generation process. Instead of conditioning on the pure geometry latent, we fine-tune our albedo generation model to synthesize roughness-metallic latents conditioned on the \textit{albedo latent} $z_{\text{albedo}}$ (either from the user input or a previously generated sample). This hierarchical approach -- i.e., geometry \textrightarrow{} albedo, then albedo \textrightarrow{} PBR -- more faithfully models the physical dependencies and leads to more plausible and detailed material generation.

For implementation, we reuse the pretrained color texture VAE by simply replacing its three color channels with roughness, metallic, and an additional zero-padding channel.
We then train the model with a similar flow matching objective $\mathcal{L}_{\text{RM}}$, additionally conditioned on the albedo latent. This design enables fully disentangled control over the final material appearance.

\subsection{{Interfacing and Interaction with \Method}}
\label{sec:method-applications}
A key advantage of our sparse latent color field is its versatility. Beyond synthesis, the learned representation serves as a powerful asset that
seamlessly integrates into existing 3D workflows and manipulated for fine-grained artistic control.

\noindent\textbf{Texture Baking for Standard Pipelines.} 
To integrate with standard rendering engines, our continuous 3D color field  $\mathcal{C}(p)$ is ``baked''~\cite{GoralTorranceGreenbergBattaile84,NishitaNakamae85} into a 2D UV texture map. Unlike multi-view projection methods~\cite{liu2024text,he2025materialmvp} that struggle to resolve inconsistent predictions, our native 3D representation makes this process straightforward and robust. Baking simply involves querying our learned function $\mathcal{C}(p)$ for each corresponding 3D point on the UV map. To produce anti-aliased textures, we supersample by averaging the colors of multiple random points within each pixel's footprint. Because our underlying field is inherently seamless, this process yields clean texture maps free of the seams and artifacts that plague projection-based methods.

\begin{table*}[htbp]
\centering
\renewcommand\arraystretch{1.0}
\renewcommand\tabcolsep{4.0pt}
\caption{\textbf{Quantitative comparison for conditional texture generation.} We evaluate \Method against leading text- and image-conditioned texture generation methods. \Method achieves superior or competitive performance across nearly all metrics, demonstrating its state-of-the-art generation quality. \textit{Lower is better for all metrics.}}
\label{tab:quantitative}

\scalebox{0.82}{
\begin{tabular}{rc|rrrrr|rrrrr} 
\toprule
\multirow{2}[1]{*}{\textbf{Method}} & \multirow{2}[1]{*}{\textbf{Condition}} & \multicolumn{5}{c|}{\textbf{Unshaded}} & \multicolumn{5}{c}{\textbf{Shaded}} \\
& & FID$\downarrow$ & $\text{FD}_\text{CLIP}\downarrow$ & $\text{FD}_\text{DINO}\downarrow$ & $\text{KD}_\text{CLIP}\downarrow$ & $\text{KD}_\text{DINO}\downarrow$  & FID$\downarrow$ & $\text{FD}_\text{CLIP}\downarrow$ & $\text{FD}_\text{DINO}\downarrow$ & $\text{KD}_\text{CLIP}\downarrow$ & $\text{KD}_\text{DINO}\downarrow$\\
\midrule

MVPaint~\cite{cheng2025mvpaint} & Text & 158.28 &  96.86 &  140.70 &  0.153 &  0.184  &  135.94 &  84.53 &  118.85 &  0.117 &  0.120\\
MaterialAnything~\cite{huang2025material} & Text &  165.50 &  96.89 &  115.72 &  0.166 &  0.129 &  126.00 &  74.76 &  108.78 &  0.090 &  0.125\\
SyncMVD~\cite{liu2024text} & Text & 126.20 &  76.25 &  88.17 &  0.072 &  0.060 & 119.38 &  66.87 &  80.99 &  0.058 &  0.047\\
TEXGen~\cite{yu2024texgen} & Text+Image &  136.58 &  86.18 &  106.66 &  0.114 &  0.094 &  130.29 &  77.62 &  97.25 &  0.088 &  0.080\\
SeqTex~\cite{yuan2025seqtex} & Text+Image &  119.31 &  61.58 &  76.14 &  0.041 &  0.033 &  116.72 & 62.20 &  74.44 &  0.051 &  0.032\\
Paint3D~\cite{zeng2024paint3d} & Text/Image&   124.73 &  69.72 &  86.47 &  0.084 &  0.059 &  116.61 &  61.70 &  80.28 &  0.063 &  0.050\\
MaterialMVP~\cite{he2025materialmvp} & Image& \cellcolor{orange!5} 113.20 & \cellcolor{orange!5} 54.79 & \cellcolor{orange!15} 67.67 & \cellcolor{orange!5} 0.048 & \cellcolor{orange!30} 0.029 &  \cellcolor{orange!30}  101.66 & \cellcolor{orange!15} 48.71 & \cellcolor{orange!15} 66.83 & \cellcolor{orange!15} 0.035 & \cellcolor{orange!15} 0.032\\
UniTEX~\cite{liang2025unitex} & Image & \cellcolor{orange!30}106.45 &  \cellcolor{orange!30}51.08 & \cellcolor{orange!15} 69.74 &  \cellcolor{orange!30}0.036 &  \cellcolor{orange!5}0.034 & \cellcolor{orange!5}105.75 &  \cellcolor{orange!5}51.62 &  \cellcolor{orange!5}69.65 &  \cellcolor{orange!5}0.038 &  \cellcolor{orange!5}0.034\\
Ours & Image & \cellcolor{orange!15} 107.46 & \cellcolor{orange!15} 53.49 & \cellcolor{orange!30} 67.16 & \cellcolor{orange!15} 0.037 & \cellcolor{orange!15} 0.032 & \cellcolor{orange!15} 101.91 & \cellcolor{orange!30} 46.28 & \cellcolor{orange!30} 64.19 & \cellcolor{orange!30} 0.026 & \cellcolor{orange!30} 0.027\\

\bottomrule
\end{tabular}
}
\end{table*}
\noindent\textbf{Local Refinement.}
Our framework naturally supports local texture alteration. Adapting techniques like RePaint~\cite{lugmayr2022repaint}, a user-defined target mesh region can be mapped to its corresponding latent voxels, which are then regenerated while keeping surrounding regions unchanged. This 3D-native local regeneration process is inherently more coherent than 2D UV repainting, as it respects the underlying geometry to ensure seamless transitions at local boundaries. This process can be utilized to achieve local texture refinement in areas where higher resolution textures are demanded. See Sec.\ref{sec:exp-applications} for more details.

\section{{Principled Data Curation}}
\label{sec:data}
Large-scale 3D datasets~\cite{deitke2023objaverse,deitke2023objaversexl} are notoriously uncurated and contain ambiguous signals. We therefore employ a principled data curation pipeline to normalize assets and ensure that our model learns from clean data.

\paragraph{Resolving Material Ambiguity.}
Public datasets contain assets with non-standard materials, such as emissive channels, that create ambiguous training signals. Our pipeline canonicalizes these materials: (1) for fully emissive surfaces, we use the emission map as the base color, as environmental lighting is irrelevant; (2) for partially emissive ones, we composite the base and emission colors with a tone mapping operator to prevent value clipping and preserve the asset's visual intent. This ensures the final color fed to our VAE remains within a valid, consistent range, preserving the visual intent of the original asset.

\paragraph{Resolving Geometric Ambiguity.}
Real-world 3D assets also contain non-standard geometry, e.g., stylized ``outline'' shells, designed for specific rendering effects. These components can corrupt the surface sampling process, as points sampled on them may be incorrectly associated with voxels of the primary object, leading to a noisy latent representation. Our curation pipeline automatically identifies and discards such geometry using heuristics based on rendering properties (e.g., backface culling, inverted normals), ensuring the VAE learns only from the intended visible surface.

\paragraph{High-Fidelity Surface Sampling.}
The fidelity of our latent representation depends on the density and quality of the input point cloud. 
To capture high-frequency textures and ensure full mesh coverage -- including small triangles and UV-distorted regions -- we employ a denser surface sampling strategy (2 million points uniformly sampled from each 3D model) to train the VAE.
For diffusion model training, we cache latents from an even denser cloud of 5 million points to provide the highest-quality latent encoding. This dense sampling creates a lossless ground truth, enabling our model to generate more fine-grained textures.

\section{Experiments}
We validate \Method through comprehensive experiments. We benchmark our core VAE representation (Sec.~\ref{sec:exp-representation}), evaluate the full generative pipeline against state-of-the-art methods (Sec.~\ref{sec:exp-generation}), and showcase downstream applications (Sec.~\ref{sec:exp-applications}).

\subsection{Implementation Details}
We train \Method on a large-scale dataset of ~1M textured 3D assets collected from Objaverse~\cite{deitke2023objaverse}, Objaverse-XL~\cite{deitke2023objaversexl}, and TexVerse~\cite{zhang2025texverse}.
Our VAE is trained for 600K steps using 16 NVIDIA A800 GPUs with a batch size of 256. The rectified flow model for albedo generation is trained for 500K steps on 32 NVIDIA A800 GPUs. We employ a progressive curriculum, first training on latents from a $64^3$ voxel grid, then fine-tuning on higher-fidelity $128^3$ latents. The PBR generator is then fine-tuned from the final albedo model by switching to a PBR material target. All models are trained using the AdamW optimizer~\cite{loshchilov2019adamw} with a constant learning rate of $1e-4$.

\begin{figure}[tbp]
    \centering
    \includegraphics[width=\linewidth,trim=00 00 00 00,clip]{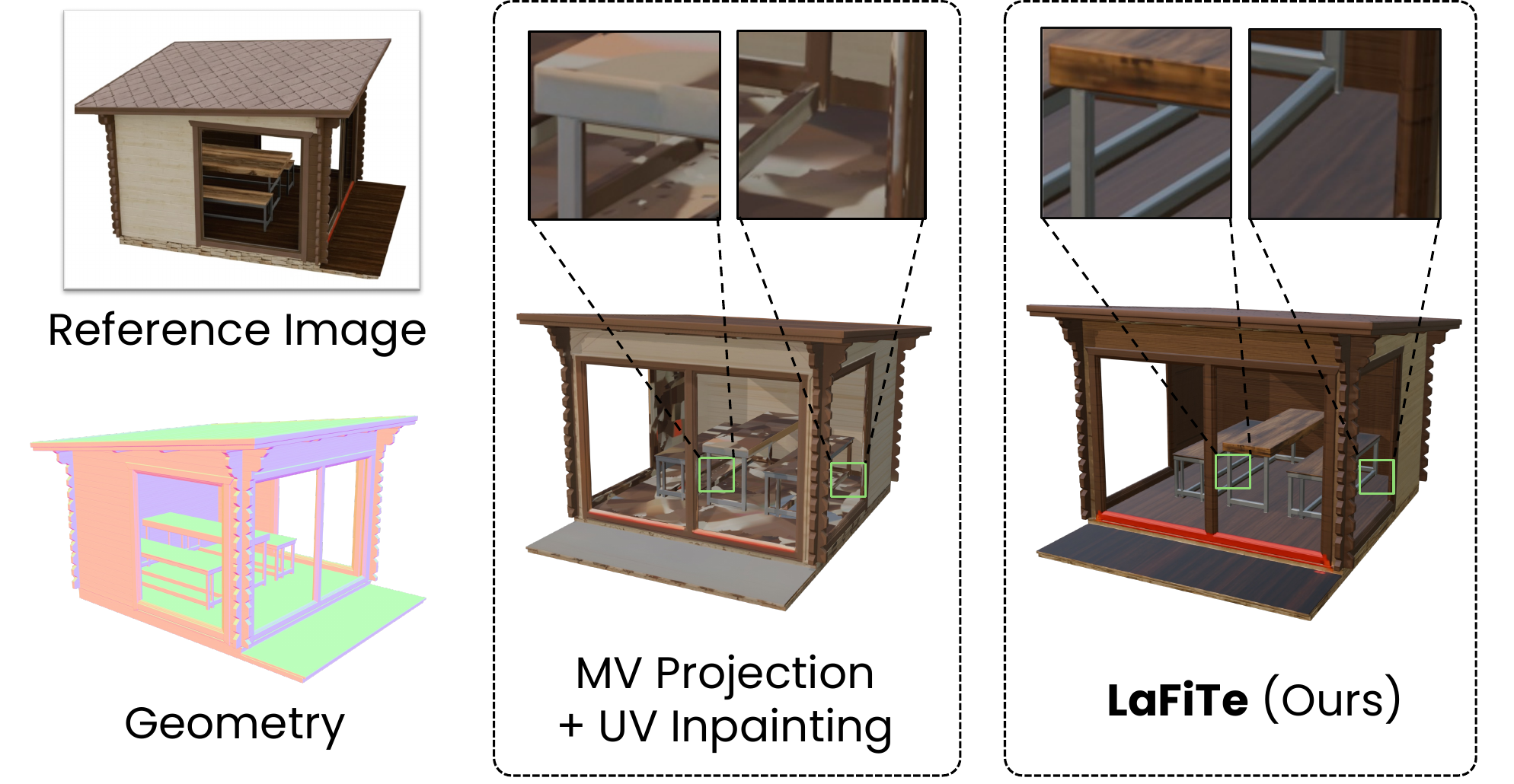}
    \caption{\textbf{Robustness to self-occlusion.} \Method's 3D-native formulation generates complete and coherent textures even in highly occluded regions where projection methods fail.}
    \label{fig:occlusion}
\end{figure}

\subsection{{Analysis of the \Method Representation}}
\label{sec:exp-representation}
We first validate the central claim of our work by evaluating the reconstruction fidelity of our VAE representation.

\noindent{\textbf{Setup.} We compare our VAE against TRELLIS~\cite{xiang2025structured}, a state-of-the-art model that represents texture by projecting pre-trained 2D DINO~\cite{oquab2023dinov2} features onto the volume as input. For a fair comparison, both models are tasked with reconstructing the original ground-truth texture from their learned latent representations. We evaluate performance using standard image-space metrics (PSNR, SSIM, LPIPS) computed across multiple rendered views.}

\begin{figure}[tbp]
    \centering
    \includegraphics[width=\linewidth,trim=00 00 00 00,clip]{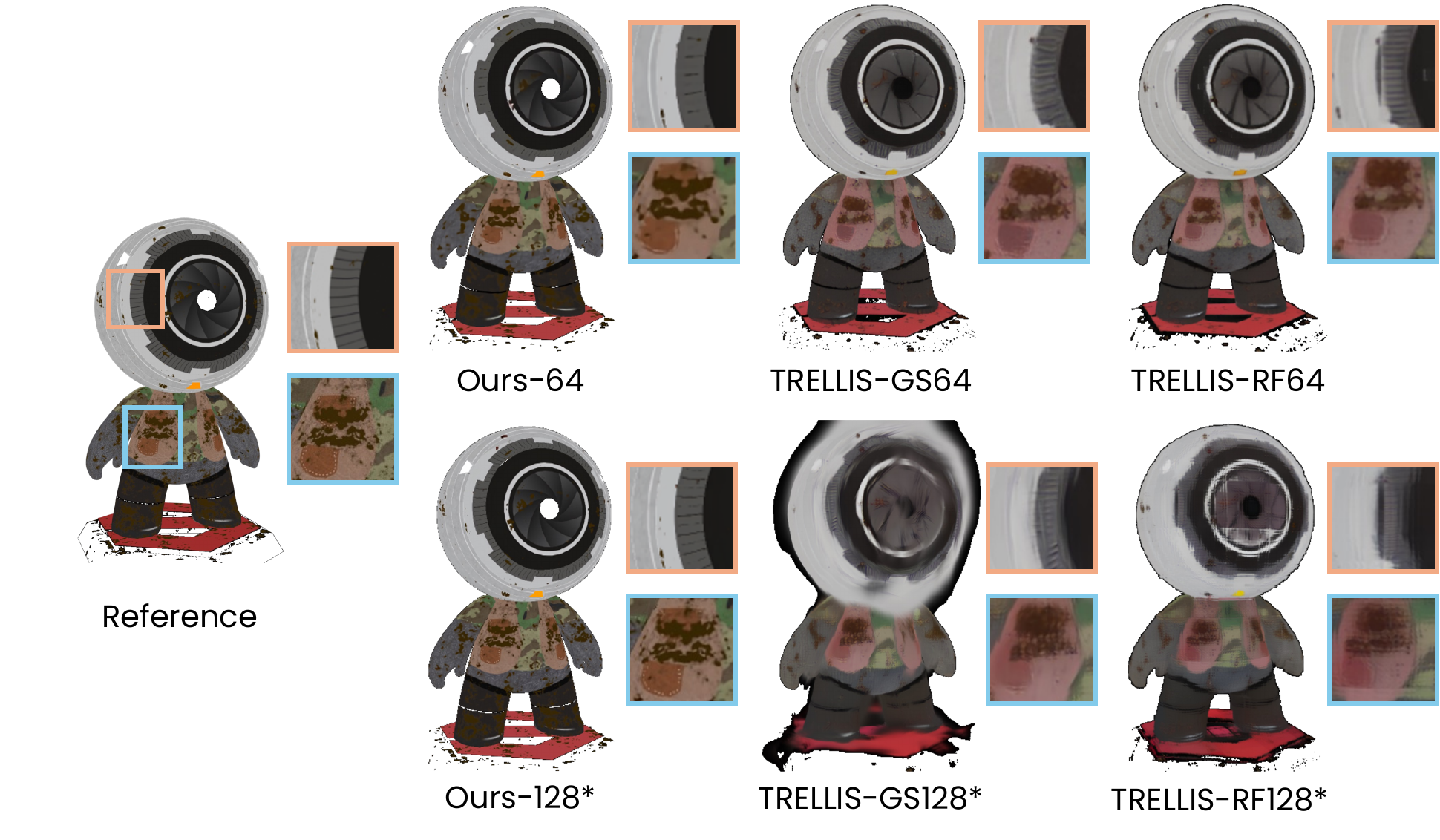}
    \caption{\textbf{Visual comparison of VAE reconstruction quality.} Our method reconstructs sharper, more detailed textures, avoiding the blurs and artifacts of the TRELLIS baseline.
    }
    \label{fig:ablation}
\end{figure}

\noindent{\textbf{Results.} As demonstrated in Tab.~\ref{tab:recon} and Fig.~\ref{fig:ablation}, \Method achieves dramatically higher reconstruction quality across all metrics and resolutions, culminating in a \emph{PSNR improvement of over 10 dB} at the $128^3$ resolution. This result provides clear evidence for our hypothesis: directly encoding a pristine 3D colored point cloud captures high-frequency surface information far more faithfully than projecting and aggregating 2D features, which inherently suffer from resolution limits, view inconsistencies, and projection ambiguities.}

\begin{table}[tbp]
\centering
\renewcommand\arraystretch{1.0}
\renewcommand\tabcolsep{8.0pt}
\caption{\textbf{VAE Reconstruction Fidelity.} Our VAE, which uses a direct point cloud input, achieves significantly higher fidelity than TRELLIS, which uses projected image features. `*' denotes the result is derived from the 64 resolution model without training on higher resolutions.}
\label{tab:recon}
\scalebox{0.95}{
\begin{tabular}{cccc} 
\toprule
\textbf{Method} & PSNR$\uparrow$ & SSIM$\uparrow$ & LPIPS$\downarrow$\\
\midrule

TRELLIS-GS64 & 21.37 & 0.833 & 0.116\\
TRELLIS-RF64 & 21.83 & 0.864 & 0.108\\
TRELLIS-GS128* & 19.28 & 0.787 & 0.195\\
TRELLIS-RF128* & 23.07 & 0.880 & 0.127\\
Ours-64 & 33.57 & 0.960 & 0.053\\
Ours-128* & 34.48 & 0.965 & 0.041\\
Ours-128 & $\textbf{34.62}$ & $\textbf{0.967}$ & $\textbf{0.039}$\\

\bottomrule
\end{tabular}
}

\end{table}

\noindent{\textbf{Scalability with Sampling Density.} We further analyze the scalability of our representation with respect to the density of the input point cloud. As shown in Tab.~\ref{tab:sample}, reconstruction quality improves consistently with denser sampling. This confirms our model's capacity to leverage richer input data and validates our dense sampling strategy (Sec.~\ref{sec:data}) as a critical component for achieving maximum fidelity.}

\begin{table}[tbp]
\centering
\renewcommand\arraystretch{1.0}
\renewcommand\tabcolsep{8.0pt}

\caption{\textbf{Effect of Sample Point Density for Reconstruction.} VAE reconstruction quality consistently improves with denser input point cloud sampling, showcasing the model's scalability.}
\label{tab:sample}
\scalebox{0.95}{
\begin{tabular}{rr|rrr} 
\toprule
\textbf{Resolution} & \textbf{\texttt{\#} of Points} & PSNR$\uparrow$ & SSIM$\uparrow$ & LPIPS$\downarrow$\\
\midrule

64 & 20,480 & 27.56 & 0.904 & 0.124 \\
64 & 204,800 & 32.22 & 0.951 & 0.067 \\
64 & 2,048,000 & 33.49 & 0.959 & 0.053 \\
128 & 20,480 & 26.07 & 0.870 & 0.141 \\
128 & 204,800 & 31.60 & 0.943 & 0.084 \\
128 & 2,048,000 & 34.30 & 0.964 & 0.043 \\
128 & 4,096,000 & $\textbf{34.45}$ & $\textbf{0.965}$ & $\textbf{0.041}$ \\

\bottomrule
\end{tabular}
}
\end{table}

\subsection{State-of-the-Art Conditional 3D Texture Generation}
\label{sec:exp-generation}

Having validated our core representation, we now evaluate the full \Method pipeline on the end-to-end task of conditional texture generation, comparing against both image-conditioned and text-conditioned state-of-the-art baselines.

\noindent{\textbf{Experimental Setup.} We curate a challenging test set designed to reflect a practical asset creation workflow. We collect approximately 800 diverse 3D meshes generated by commercial AI tools~\cite{tripo3d_ai,hunyuan3d_tencent,hyper3d_ai,meshy_ai,hitem3d_ai}, using a set of 200 prompt images from various sources~\cite{ebert20253d,xiang2025structured,wu2025direct3d}. This ensures the evaluation is performed on novel, ``in-the-wild'' geometries rather than familiar assets from open-source training sets. For text-based methods, we use a powerful VLM (Qwen-2.5VL~\cite{Qwen2.5-VL}) to generate prompts from the reference images, ensuring a fair comparison.}

\begin{figure*}[tbp]
    \centering
    \includegraphics[width=\linewidth,trim=00 00 00 00,clip]{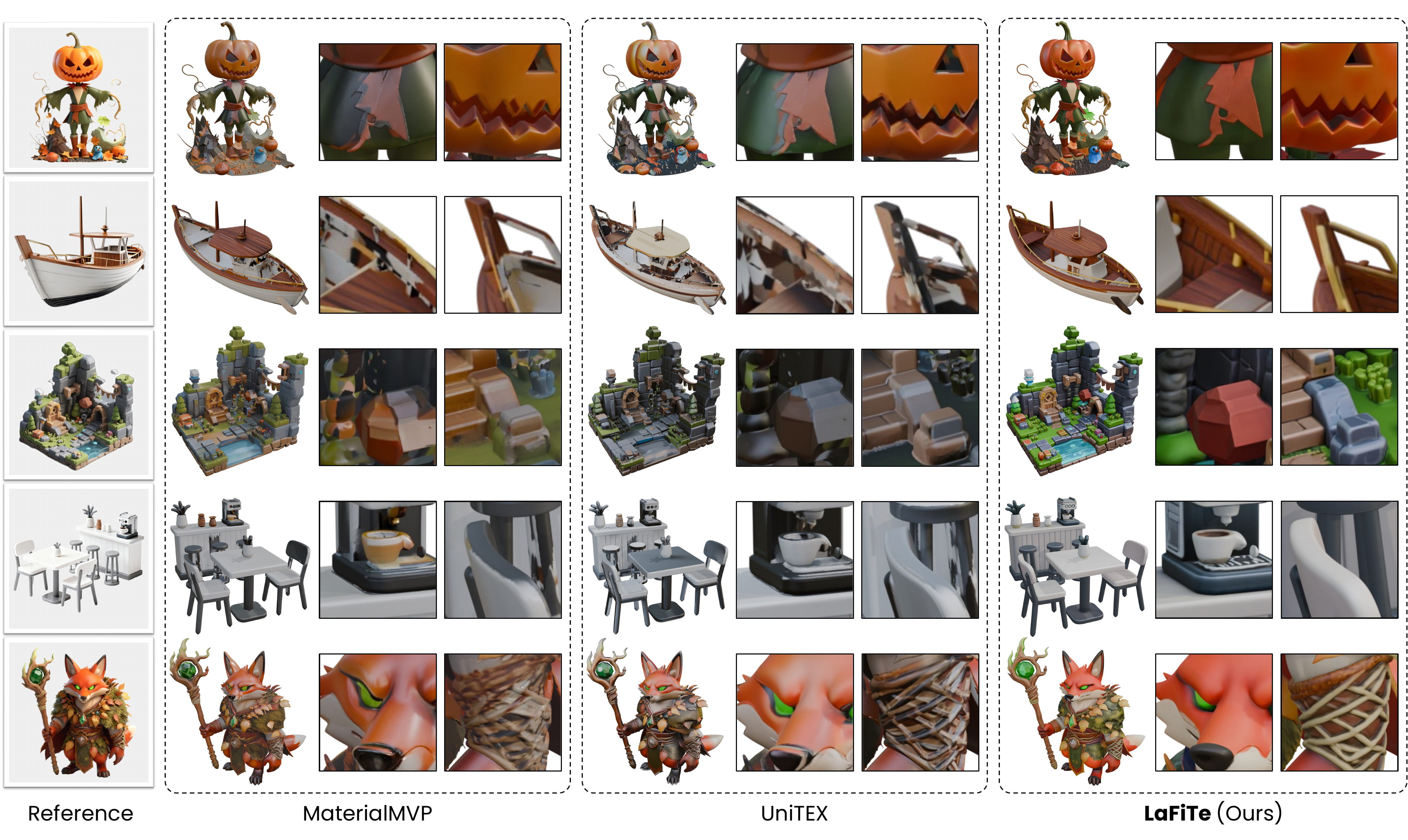}
    \caption{\textbf{Qualitative comparison of image-conditioned 3D texture generation.} Textures generated by \Method are more faithfully aligned to both the reference image and the given geometry, and are free from seams or inconsistencies.}
    \label{fig:qualitative_image}
\end{figure*}

\noindent\textbf{Metrics and Evaluation.} We render all generated texturing results from three canonical views under both unshaded and shaded settings. For methods that do not generate PBR channels, shaded results are rendered with a default material (metallic=0, roughness=0.5). To measure the fidelity and alignment with the input condition, we compute a suite of standard generative modeling metrics: Fréchet Distance (FD)~\cite{alt1995frechet} and Kernel Distance (KD)~\cite{phillips2011kernel} using feature spaces from CLIP~\cite{radford2021clip} and DINO~\cite{oquab2023dinov2} to capture semantic and stylistic similarity.

\noindent\textbf{Quantitative Results.} As shown in Table~\ref{tab:quantitative}, \Method demonstrates superior performance across nearly all metrics and conditions. In the image-conditioned task, \Method achieves the best or second-best scores on all metrics, significantly outperforming prior work in $\text{FD}_{\text{CLIP}}$ and $\text{KD}_{\text{DINO}}$ under the \textit{shaded} setting. This confirms that our framework, which leverages a high-fidelity representation and precise 3D geometric conditioning, produces textures that are more faithful to the input prompt than any alternative method.

\noindent{\textbf{Qualitative Results.} Beyond metrics, \Method exhibits clear qualitative advantages, which we analyze in terms of \textbf{Integrity}, \textbf{Continuity}, and \textbf{Versatility}. As visualized in Fig.~\ref{fig:qualitative_image}, our method consistently generates textures that fully cover the entire mesh (\textbf{Integrity}), are free from UV seams or projection artifacts (\textbf{Continuity}), and generalize well to a wide variety of object categories (\textbf{Versatility}). In contrast, competing methods often produce visible seams, blurred details, or incomplete textures, particularly on complex or self-occluded geometry.}

\begin{figure}[tbp]
    \centering
    \includegraphics[width=\linewidth,trim=0 0 0 0,clip]{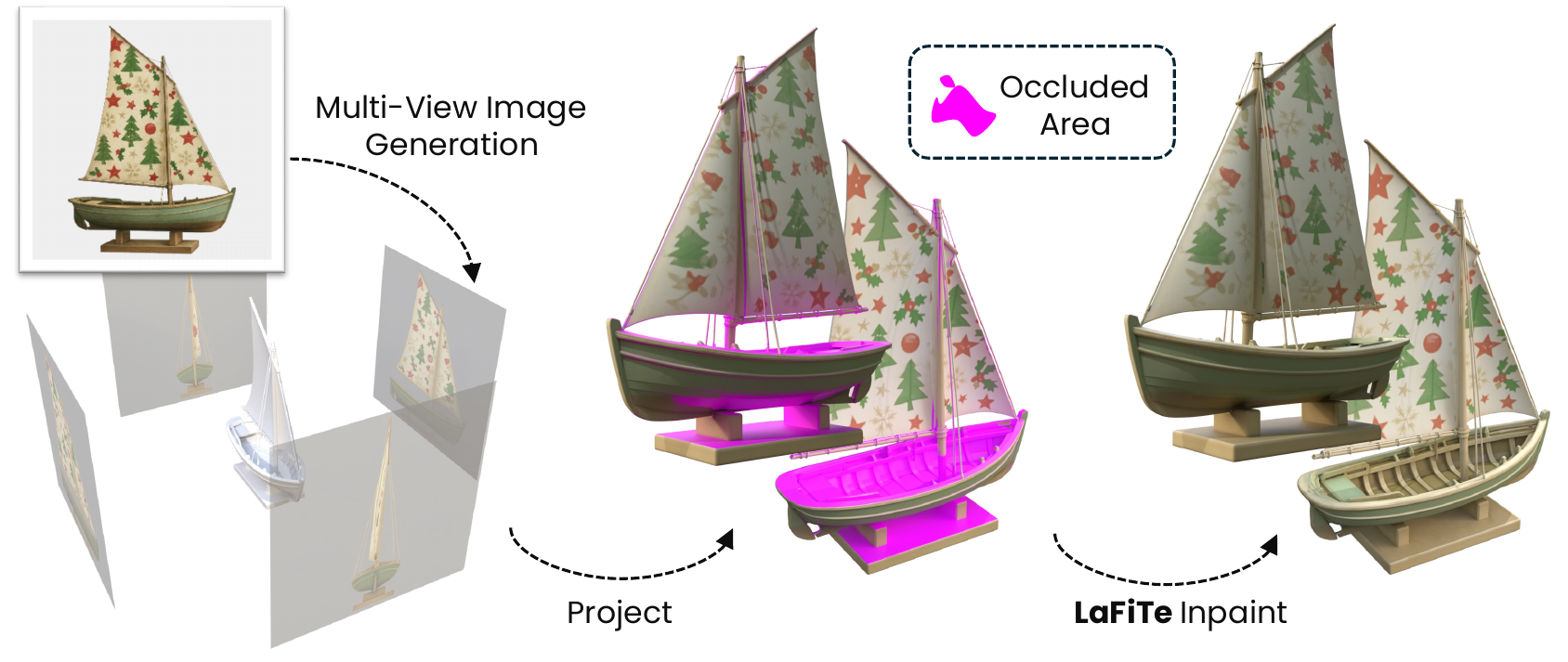}
    \vspace{-6mm}
    \caption{\textbf{Integration with multi-view projections.} \Method can be used to complete occluded regions for the partial 3D texture projected from multi-view images.}
    \label{fig:inpainting}
    \vspace{-3mm}
\end{figure}

\begin{figure}[tbp]
    \centering
    \includegraphics[width=0.95\linewidth,trim=0 40 200 0,clip]{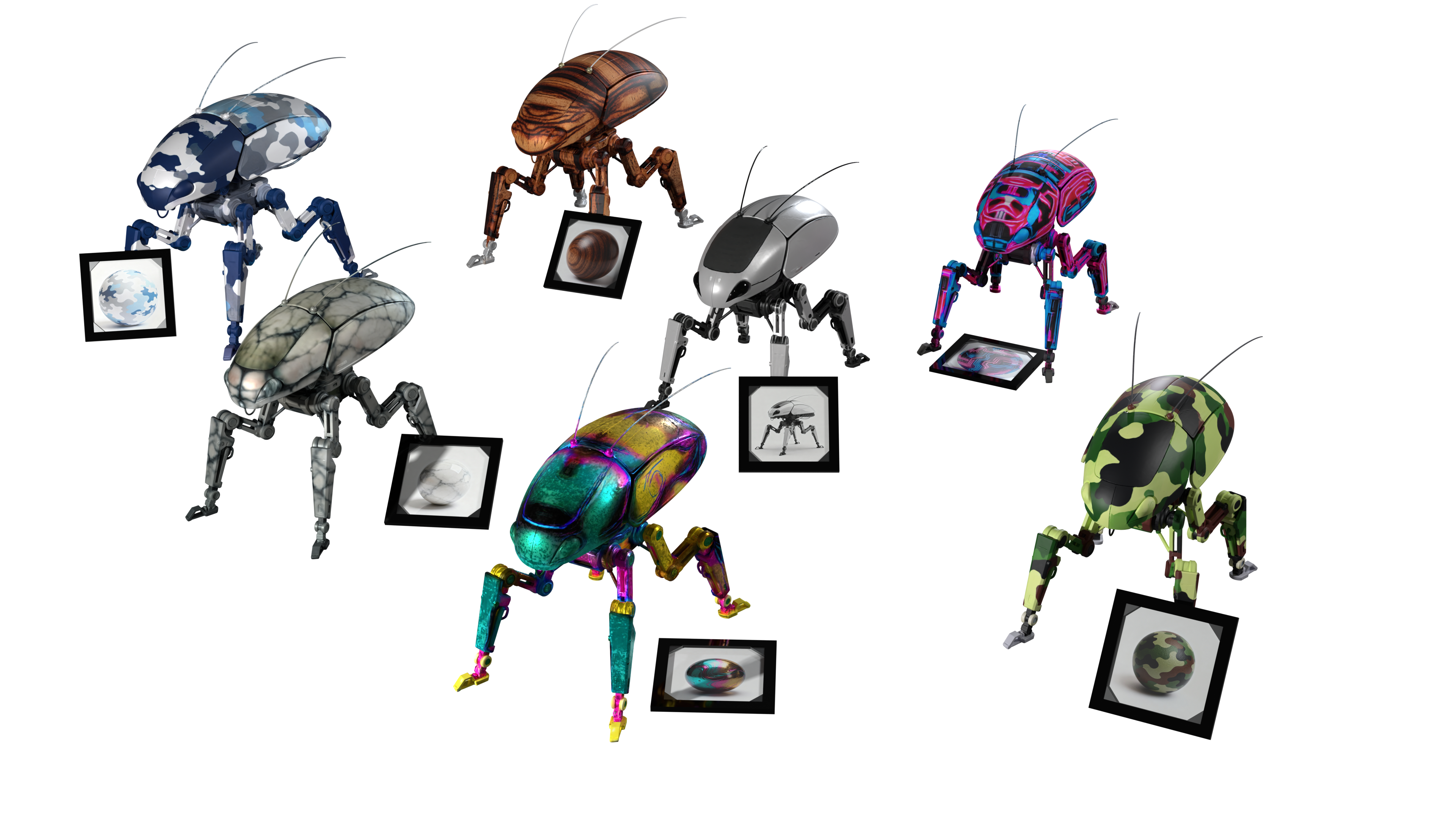}
    \vspace{-2mm}
    \caption{\textbf{Material Transfer.} \Method can generate textures with PBR materials conditioned on a 2D material image.}
    \vspace{-6mm}
    \label{fig:material}
\end{figure}

\begin{figure}[tbp]
    \centering
    \includegraphics[width=1.0\linewidth,trim=0 0 0 0,clip]{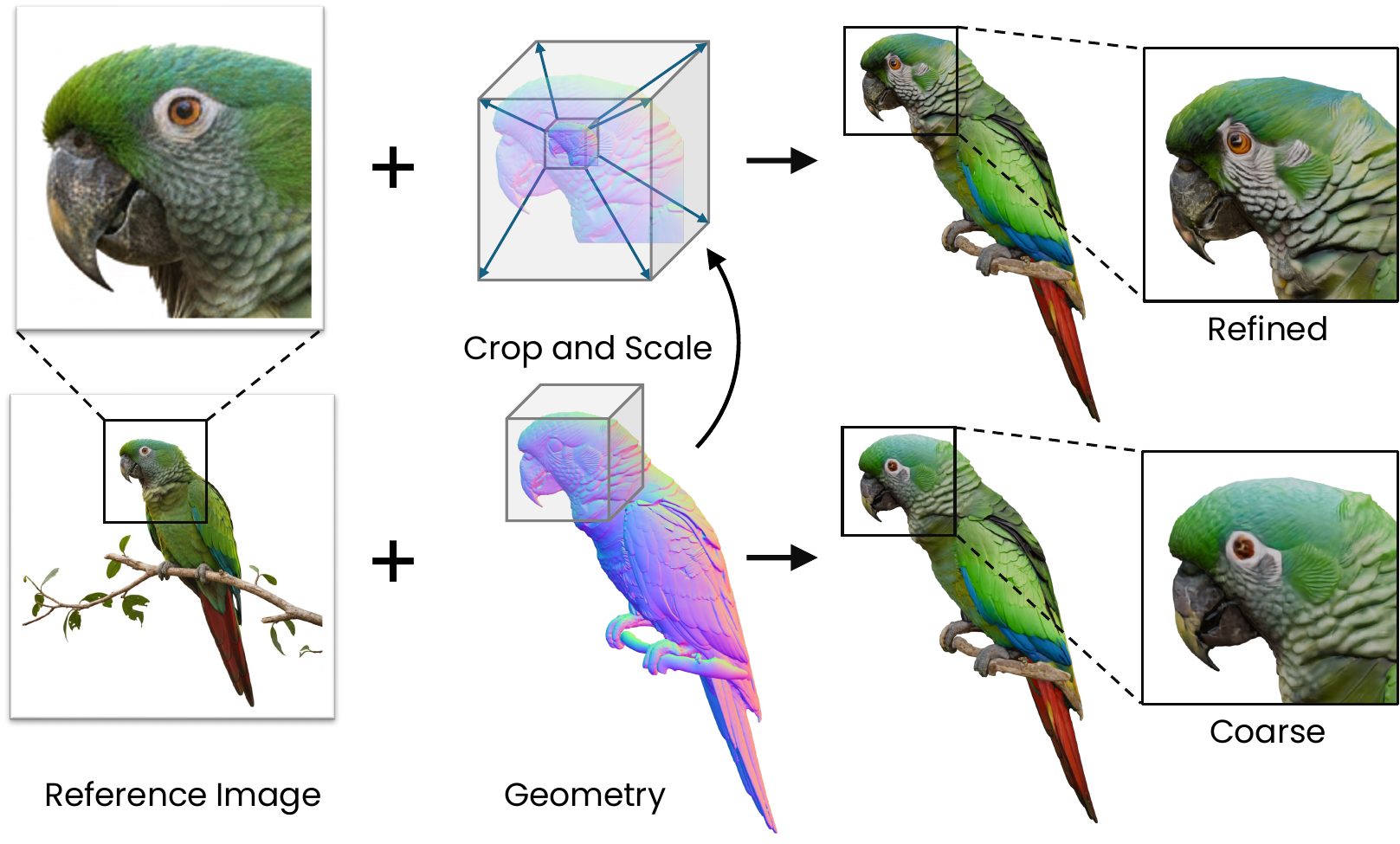}
    \caption{\textbf{Local Refinement.} \Method can further enhance the generation quality by refining a local region.}
    \label{fig:refinement}
\end{figure}

\subsection{Downstream Applications}
\label{sec:exp-applications}
A key advantage of our learned latent field is its flexibility. We demonstrate its power through several applications:
\begin{itemize}
    \item \textbf{Integration with Multi-View Pipelines:} The structured latent representation enables \Method to be easily integrated with multi-view projections by inpainting the unseen area. Fig.~\ref{fig:inpainting} shows that our method can robustly fill occluded regions while keeping the projected areas unchanged.
    \item \textbf{Material Transfer:} Fig.~\ref{fig:material} demonstrates direct material transfer, where a 2D material image is used as the image condition to synthesize seamless PBR textures on the mesh.
    \item \textbf{Local Refinement:} Fig.~\ref{fig:refinement} shows an example of local texture refinement using our method, resulting in a much higher level of local details.
\end{itemize}

\section{Conclusion}

We introduced \Method, a generative framework that fundamentally advances native 3D texturing. We identified the critical bottleneck as the lack of a powerful, topology-agnostic representation and solved it by learning to generate textures as a 3D structured latent color field. Our VAE-based representation achieves state-of-the-art fidelity, and our unified design for geometric conditioning ensures seamless, high-quality generation. By solving this core representation problem, \Method significantly outperforms existing methods and provides a robust foundation for future 3D generative workflows.

\noindent\textbf{Limitations and Future Work.}
Lafite's 3D-native approach lacks the priors of large-scale 2D generation models compared to multi-view-based 3D texturing methods, limiting its ability on semantic soundness and text rendering quality. Future work may explore leveraging the vast prior knowledge embedded in large-scale image and video datasets to mitigate the scarcity of high-quality textured 3D training data.


{
    \small
    \bibliographystyle{ieeenat_fullname}
    \bibliography{main}
}

\end{document}